\documentclass[Journal,letterpaper,NoLists,InsideFigs,SingleSpace,NoLineNumbers]{ascelike-new}
\usepackage[utf8]{inputenc}
\usepackage[T1]{fontenc}
\usepackage{lmodern}
\usepackage{graphicx}
\usepackage[style=base,figurename=Fig.,labelfont=bf,labelsep=period]{caption}
\usepackage{amsmath}
\usepackage{newtxtext,newtxmath}
\usepackage[colorlinks=true,citecolor=red,linkcolor=black]{hyperref}
\usepackage{booktabs}
\usepackage{multirow}
\usepackage{array}
\usepackage{tabularx}
\usepackage{float}
\usepackage{lineno} 

\begin{document}
\title{Vision Transformer-Based Multi-Level Feature Fusion for Multi-Label Sewer Defect Classification}

\author[1]{Xu Fang}
\author[2]{Zhuoran Wang}
\author[3]{Qing Li}
\author[1]{Shengyu Zhang}
\author[2]{Guanzhi Deng}
\author[1]{Jianbiao He}
\author[4]{Qingquan Li}

\affil[1]{Shenzhen Polytechnic University, Shenzhen, China. E-mails: fangxu.hello@gmail.com; zsyhcxs@szpu.edu.cn. Corresponding author: Jianbiao He; E-mail: hejianbiao@szpu.edu.cn}
\affil[2]{City University of Hong Kong, Hong Kong, China. E-mails: wangzr040220@gmail.com; guanzdeng2-c@my.cityu.edu.hk}
\affil[3]{Pengcheng Laboratory, Shenzhen, China. E-mail: lqing900205@gmail.com}
\affil[4]{Shenzhen University, Shenzhen, China. E-mail: liqq@szu.edu.cn}

\maketitle

\begin{abstract}
Automated classification of sewer defects is essential for infrastructure condition assessment and maintenance decision-making, but existing deep learning methods struggle to balance classification accuracy and computational complexity in large-scale multi-label scenarios. This study develops Sewer-Transformer-ML, a hierarchical vision Transformer with multi-level feature fusion, together with two lightweight architectures, Sewer-MobileNet-ML and Sewer-Mobile-TransNet, for resource-constrained inspection scenarios. On the Sewer-ML test set, Sewer-Transformer-ML-Base achieved an $F2_{\text{CIW}}$ of 65.68\% and an $F1_{\text{Normal}}$ of 92.68\%, ranking first on the public leaderboard and exceeding the second-ranked method by 7.6 percentage points in $F2_{\text{CIW}}$. Sewer-MobileNet-ML achieved an $F2_{\text{CIW}}$ of 65.73\% with only 17 M parameters, representing an approximately 95\% parameter reduction relative to the base model. Under the standard Sewer-Capsule data split, Sewer-Mobile-TransNet achieved 96.43\% classification accuracy. When the training set was reduced to 1,177 images, pretraining on Sewer-ML consistently improved model performance. Ablation experiments further showed that direct concatenation was more effective for Transformer features, whereas attention-based fusion better supported multiscale CNN features. These findings provide a computational basis for automated sewer inspection, lightweight model design, and adaptation across civil infrastructure inspection platforms.
\end{abstract}

\section{Practical Applications}\label{sec:practical}

Municipal sewer inspections generate large volumes of closed-circuit television images and videos that are commonly reviewed manually, making the process time-consuming and dependent on inspector experience and workload. The proposed models can support this workflow by automatically screening inspection images, distinguishing normal conditions from different defect types, and assigning multiple labels when several defects occur in the same image. The resulting classifications can help engineers prioritize images or pipe segments for further examination and provide supporting information for condition assessment, maintenance prioritization, and rehabilitation planning. Because the lightweight models substantially reduce the number of parameters while retaining high classification accuracy, they have the potential to be integrated into sewer inspection robots, embedded computing devices, or field video-processing systems. The cross-dataset experiments also indicate that knowledge learned from large CCTV data sets can improve adaptation to images collected by emerging sewer-capsule platforms, reducing the amount of labeled data required when introducing a new inspection device. These results demonstrate technical application potential; however, inference speed, energy consumption, and long-term reliability on operational field hardware remain to be validated.

\section{Introduction}\label{sec:intro}

Accurate inspection and quantitative assessment of urban underground drainage systems are critical components of municipal infrastructure health monitoring, essential for ensuring urban environmental safety and resilient operation in the context of smart city development and sustainable urbanization. As subsurface assets that form the backbone of modern urban areas, these pipeline networks require continuous condition monitoring for public health protection and flood prevention, where failures can lead to significant economic losses and social disruption. However, pipelines are susceptible to natural environmental factors and engineering construction impacts during long-term service, developing various defects such as cracking, displacement, corrosion, or blockage, which may cause safety accidents like water leakage, ground subsidence, or even collapse. Therefore, automated inspection and intelligent analysis of drainage pipelines are necessary to avoid pipeline performance deterioration or sudden structural failure, establishing a quantitative basis for intelligent infrastructure management in modern cities.

Although deep learning models have achieved success in speech recognition, face recognition and other fields relying on large-scale datasets like ImageNet and MSCOCO, they still face three major challenges in automated multi-label classification of sewer defect images from closed-circuit television (CCTV) and robotic inspections for large-scale municipal asset management: (1) Environmental complexity: insufficient lighting and high noise in internal pipeline imaging under complex field conditions, diverse defect morphologies and significant scale variations, leading to insufficient classification accuracy of existing models; (2) Computational resource constraints: deep models have large parameter counts, making edge deployment on mobile or embedded devices difficult; (3) Data and evaluation bottlenecks: commercial restrictions result in a lack of publicly available large-scale benchmark datasets, and emerging robotic collection methods like sewer pipeline capsule systems produce low-quality data with few samples, making it difficult to reproduce and fairly compare existing methods.

To address the above challenges in infrastructure condition assessment, the main contributions of this paper include:
\begin{enumerate}
\item Proposing the Sewer-Transformer-ML model as a novel intelligent inspection framework: Based on a hierarchical vision Transformer architecture, it enhances defect representation capability by integrating multi-level features, achieving first place on the largest open-source benchmark Sewer-ML (establishing a new classification benchmark with $F2_{\text{CIW}}$ 65.68\%, $F1_{\text{Normal}}$ 92.68\%), significantly outperforming the second-best method by 7.6 percentage points (65.68\% vs. 58.08\%).
\item Designing lightweight solutions: Constructing Sewer-MobileNet-ML based on MobileNetV3 and Sewer-Mobile-TransNet, achieving a good trade-off between accuracy and complexity with approximately 95\% parameter reduction while maintaining state-of-the-art accuracy (65.73\% $F2_{\text{CIW}}$,  $F1_{\text{Normal}}$ 89.53\%), validating that small models can achieve superior performance for automated field inspection.
\item Revealing feature fusion principles: Through comprehensive and systematic comparison, we reveal that direct stacking of Transformer multi-level features achieves optimal results, while CNN multi-scale features are more suitable for fusion through multi-head attention mechanisms, providing scientific basis for deep learning network structure design in infrastructure defect detection.
\item Validating cross-domain transferability: Through comprehensive transfer learning experiments, we demonstrate that transferring pre-trained knowledge from large-scale CCTV inspection datasets to the small-scale Sewer-Capsule Dataset collected by novel robotic systems significantly improves classification performance. Sewer-Mobile-TransNet achieved 96.43\% accuracy under the standard data split, and pretraining consistently improved model performance when the training set was reduced to 1,177 images, validating cross-domain transferability and enhancing practicality for municipal infrastructure maintenance.
\end{enumerate}

\section{Related Work}\label{sec:related}

Previous automated sewer inspection methods can be broadly grouped into traditional image-processing and machine-learning approaches and more recent deep-learning approaches.

\subsection{Traditional Image Processing and Machine Learning Methods for Pipeline Defect Detection}

Early systems combined hand-crafted texture, shape, or edge features with neural-network, support-vector-machine, or random-forest classifiers \cite{yang2008automated,shehab2005automated,myrans2019automated}. Other pipelines used edge detection and morphological processing to identify visually distinctive defects such as cracks and open joints \cite{su2014application,halfawy2014efficient}. Although these approaches reduced part of the manual inspection workload, their multi-stage processing and task-specific feature design limited generalization across variable illumination, noise, pipe materials, and defect morphologies.

\subsection{Deep Learning Methods for Pipeline Defect Detection}

CNNs subsequently enabled end-to-end feature learning for sewer defect classification. Early multi-label systems used multiple binary subnetworks or shallow CNNs \cite{kumar2018automated,meijer2019defect}, while hierarchical approaches separated defect screening from class prediction to address imbalance and task complexity \cite{li2019sewer,xie2019automatic}. The release of Sewer-ML established a large-scale multi-label benchmark with class-importance-weighted evaluation, enabling fairer comparison of these methods \cite{haurum2021sewer}.

Recent research has examined stronger CNN backbones, vision Transformers, and hybrid CNN-Transformer architectures. Cross-domain benchmarking indicates that pretrained CNNs can remain competitive on small target datasets \cite{meng2025enhancing}, whereas hybrid models combine local convolutional features with long-range attention \cite{goharinezhad2025automated,zhang2024weak}. Nevertheless, existing studies provide limited systematic evidence on how multi-level Transformer features and multiscale CNN features should be fused for large-scale multi-label sewer classification. The balance between classification performance and lightweight deployment, as well as transfer from large CCTV datasets to data collected by emerging robotic platforms, also remains insufficiently studied. Accordingly, this work compares pure Transformer, lightweight CNN, and hybrid architectures and evaluates their feature-fusion and cross-domain transfer behavior \cite{liu2021swin,dosovitskiy2021image,nguyen2025sewer}.

\section{Methodology}\label{sec:method}

To address the multi-label classification challenge in sewer pipeline defect detection, this paper proposes a hierarchical vision Transformer-based inspection framework with Swin Transformer as the backbone. The main model, Sewer-Transformer-ML, achieves high recognition accuracy through multi-level feature fusion. To balance performance and computational efficiency, two lightweight variants are developed: Sewer-MobileNet-ML (pure CNN) and Sewer-Mobile-TransNet (hybrid CNN-Transformer). Furthermore, a Category Importance Weighting loss is introduced to mitigate class imbalance.

\subsection{Multi-Label Classification Formulation}

Unlike single-label tasks where each image belongs to exactly one class, sewer inspection images may contain multiple co-occurring defects, necessitating multi-label classification. While the overall architecture (Backbone, Neck, Head) remains similar, the key distinction lies in the output encoding and loss function.

Multi-label classification treats each defect class as an independent binary task. The model outputs $\mathbf{x} \in \mathbb{R}^C$ are passed through element-wise Sigmoid activation, and Binary Cross-Entropy (BCE) loss is applied:

\begin{linenomath*}
\begin{equation}
\label{eq:bce}
L(x, y) = \frac{1}{C} \sum_{i=1}^{C} - \Bigl[ p_i y_i \log(\sigma(x_i)) + (1 - y_i) \log(1 - \sigma(x_i)) \Bigr]
\end{equation}
\end{linenomath*}

where $C$ is the number of defect classes, $\mathbf{y} \in \{0,1\}^C$ is the ground-truth label vector, $\sigma(\cdot)$ is the Sigmoid function, and $p_i$ is a class-specific weighting factor to address dataset imbalance.

\subsection{Sewer-Transformer-ML: Multi-level Vision Transformer}

\subsubsection{Sewer-Transformer-ML Network Structure}

\begin{figure}[htbp]
    \centering
    \includegraphics[width=0.7\linewidth]{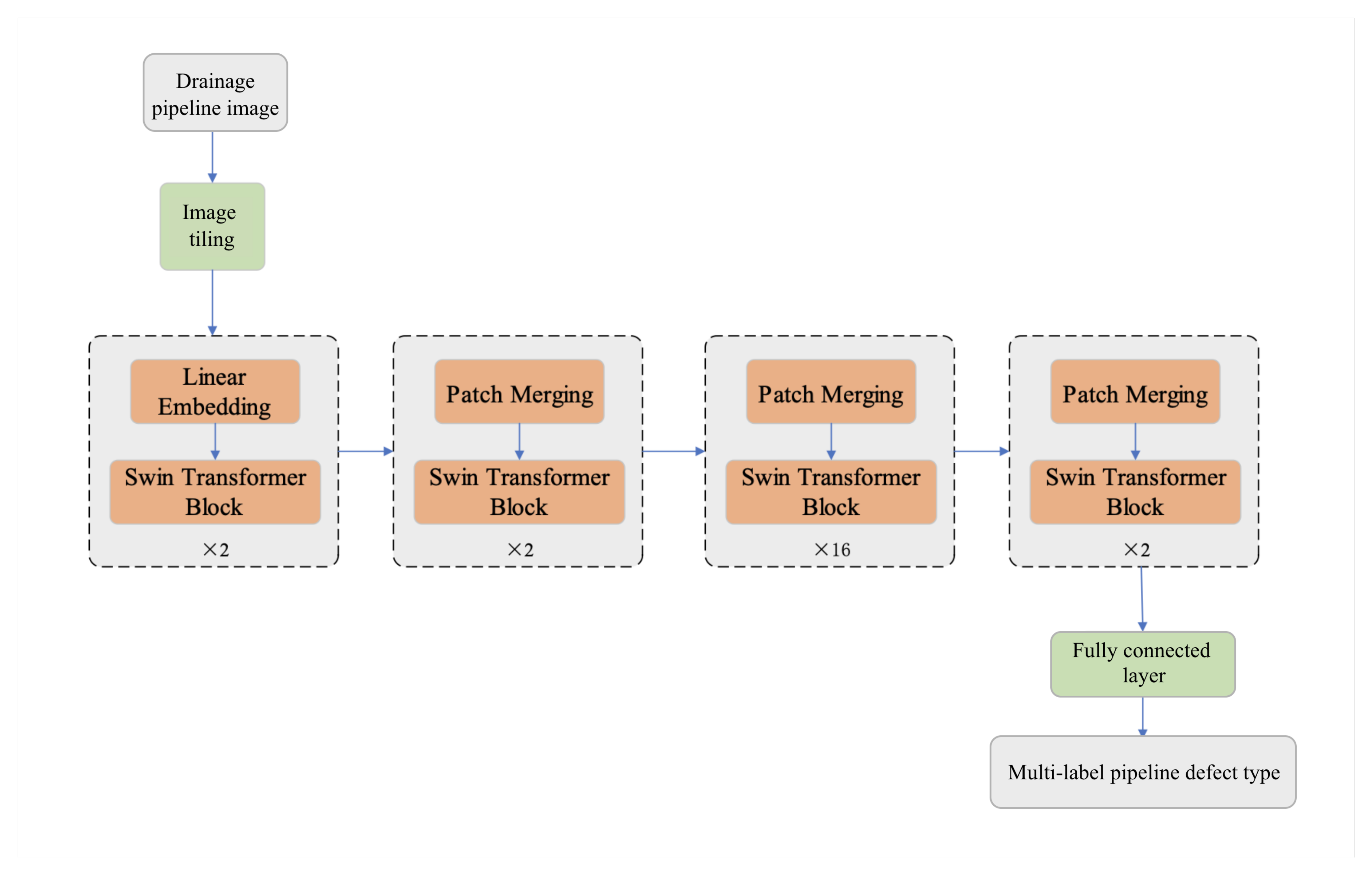}
    \caption{Multi-label sewer defect classification model based on a multi-level vision Transformer: Sewer-Transformer-ML (Base version)}
    \label{fig:7}
\end{figure}

To improve the accuracy and practicality of sewer pipeline defect recognition, this section proposes a multi-label classification model based on a multi-level vision Transformer, namely Sewer-Transformer-ML. This model uses Swin Transformer\cite{liu2021swin} as the backbone network, eliminating convolutional operations in the backbone and instead extracting defect features through self-attention mechanisms and fusing multi-level semantic information to enhance feature representation capability. The overall architecture is shown in Figure\ref{fig:7}, and is divided into Tiny, Small, Base, and Large versions based on the number of Swin Transformer Blocks in Stage 3, with Block counts of 12, 12, 16, and 24 respectively.

Sewer-Transformer-ML adopts a hierarchical design. First, the Patch Partition module divides the input image into non-overlapping patches. Setting patch size as $4 \times 4$, each patch is flattened to a 48-dimensional feature vector ($4 \times 4 \times 3$). Subsequently, the Linear Embedding layer maps it to dimension $C$, with corresponding $C$ values of 96, 96, 128, and 192 for the four versions.

Multi-level feature representation draws on the idea of multi-scale feature extraction in convolutional neural networks. The entire network consists of multiple feature extraction stages (Stage), finally connecting to fully connected layers after feature fusion. As network depth increases, the Patch Merge module gradually reduces feature map resolution to decrease computational complexity, while each Swin Transformer Block performs modeling within local windows. The feature map dimensions evolve as follows:
\begin{itemize}
    \item Stage 1: Maintained as $(H/4, W/4, C)$
    \item Stage 2: Downsampled to $(H/8, W/8, 2C)$
    \item Stage 3: Downsampled to $(H/16, W/16, 4C)$
    \item Stage 4: Downsampled to $(H/16, W/16, 8C)$
\end{itemize}
This hierarchical design produces multi-scale features similar to ResNet and MobileNet, enabling flexible adaptation to diverse computer vision tasks.

Following the original Swin Transformer formulation \cite{liu2021swin}, each stage alternates window-based multi-head self-attention (W-MSA) and shifted-window multi-head self-attention (SW-MSA). W-MSA limits attention computation to local windows, while SW-MSA enables information exchange across adjacent windows. This hierarchical local-attention design provides computationally efficient modeling of defect patterns at different spatial scales; the present study focuses on how features from these stages are fused for multi-label sewer defect classification rather than modifying the standard Swin block.

\subsubsection{Multi-level Vision Transformer Feature Fusion}

\begin{figure}[htbp]
    \centering
    \includegraphics[width=0.5\linewidth]{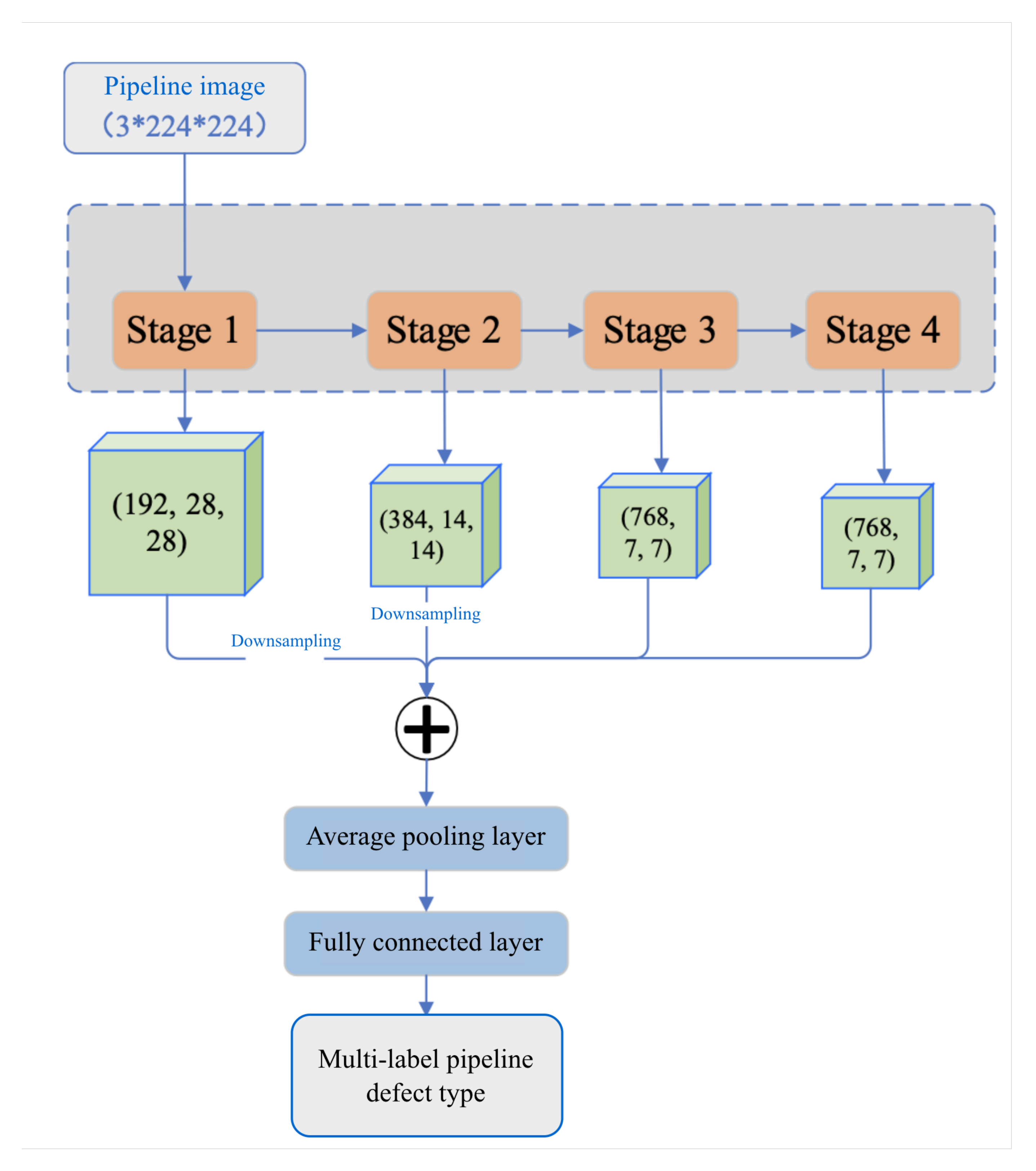}
    \caption{Multi-label sewer defect classification based on a vision Transformer and multi-level feature fusion}
    \label{10}
\end{figure}

The hierarchical feature representation in vision Transformer structures has inherent similarity with multi-scale feature extraction in CNN. In CNN, fusing features at different scales (i.e., high-level semantics and low-level details) has significantly improved model performance in various visual task network structure designs. Inspired by this, we propose fusing multi-level features of Transformer to enhance representation capability for pipeline defects with complex and variable morphologies. The overall architecture is shown in Figure \ref{10}.
To fully fuse multi-level features extracted by Swin Transformer Blocks at different stages and prevent information loss from continuous downsampling, low-level features are downsampled to the same size as high-level features and then concatenated before being fed into fully connected layers for multi-label classification. Meanwhile, to explore the characteristics of CNN multi-scale features and Transformer multi-level features, this study compares the following two attention fusion strategies:
\begin{enumerate}
    \item Multi-scale feature fusion based on separated attention mechanism structure units \cite{zhang2022resnest};
    \item Convolutional module based on channel-spatial attention \cite{woo2018cbam}.
\end{enumerate}

Specifically, after concatenating multi-level features upsampled or downsampled to uniform size, they are input into fusion units for attention screening to obtain recalibrated features (emphasizing important features and suppressing unimportant ones), which are then fed into fully connected layers for classification.

\subsection{Lightweight Variants for Edge Deployment}

To reduce deployment cost, the two lightweight variants use MobileNetV3 \cite{howard2019searching}, whose standard inverted residual bottlenecks combine depthwise separable convolution, channel expansion and projection, and squeeze-and-excitation. These established components are adopted without structural modification; the distinction between the variants lies in whether the extracted multi-scale CNN features are classified directly or fused using shifted-window self-attention.

\subsubsection{Sewer-MobileNet-ML Network Structure}

Sewer-MobileNet-ML adapts the MobileNetV3-Large backbone to multi-label sewer defect classification by replacing its original classifier with a sigmoid-based multi-label prediction head. The resulting model contains approximately 17M parameters and serves as the pure-CNN lightweight baseline for evaluating the performance--complexity trade-off.

\subsubsection{Sewer-Mobile-TransNet Network Structure}

\begin{figure}[htbp]
    \centering
    \includegraphics[width=0.3\linewidth]{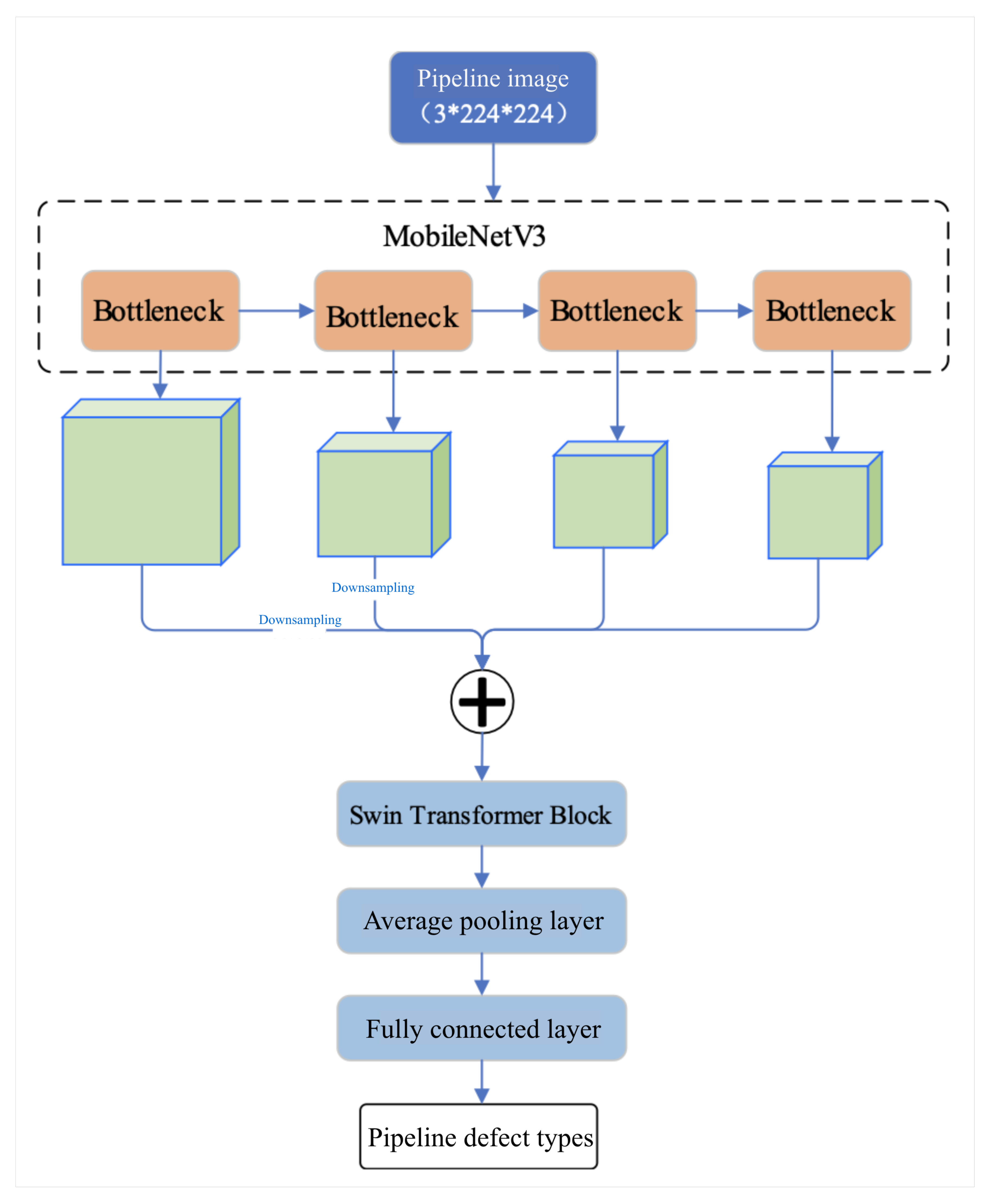}
    \caption{Sewer-Mobile-TransNet for sewer defect classification based on a lightweight network and multi-head shifted-window self-attention}
    \label{fig:16}
\end{figure}

Sewer-Mobile-TransNet is a multi-label sewer defect classification model based on lightweight networks and multi-head shifted-window self-attention, with its overall structure shown in Figure~\ref{fig:16}. This model uses MobileNetV3~\cite{howard2019searching} as the backbone, with Swin Transformer Block based on the shifted-window multi-head self-attention mechanism serving as an important module in the Sewer-Mobile-TransNet structure. Multiple cascaded Bottleneck modules extract multi-scale CNN features. To fully fuse multi-scale information, Swin Transformer Block is introduced as a feature fusion module: first, feature maps at different scales are aligned to the same size through upsampling or downsampling and concatenated, then the shifted-window multi-head self-attention mechanism is used to screen and semantically correlate the concatenated features, modeling their spatial relationships to enhance feature representation capability for pipeline defects.

\section{Datasets and Evaluation Methods}\label{sec:dataset}

\subsection{Dataset Introduction}

This section employs two types of datasets: the CCTV pipeline image dataset Sewer-ML Dataset and the pipeline capsule dataset Sewer-Capsule Dataset. The former is primarily used to validate the capability of visual attention mechanisms in pipeline defect classification tasks and provide pre-trained weights for the latter for transfer learning.

\begin{table}[htbp]
\centering
\caption{Sewer-ML Dataset Defect Category Codes and Weight Factors \protect\cite{haurum2021sewer}}
\label{tab:1}
\begin{tabular}{p{1.2cm} p{5.5cm} p{1.4cm}}
\toprule
\textbf{Code} & \textbf{Description} & \textbf{CIW} \\
\midrule
RB & Cracks, breaks, and collapses & 1.0000 \\
OB & Surface damage & 0.5518 \\
PF & Production error & 0.2896 \\
DE & Deformation & 0.1622 \\
FS & Displaced joint & 0.6419 \\
IS & Intruding sealing material & 0.1847 \\
RO & Roots & 0.3559 \\
IN & Infiltration & 0.3131 \\
AF & Settled deposits & 0.0811 \\
BE & Attached deposits & 0.2275 \\
FO & Obstacle & 0.2477 \\
GR & Branch pipe & 0.0901 \\
PH & Chiseled connection & 0.4167 \\
PB & Drilled connection & 0.4167 \\
OS & Lateral reinstatement cuts & 0.9009 \\
OP & Connection with transition profile & 0.3829 \\
OK & Construction changes & 0.4396 \\
\bottomrule
\end{tabular}
\end{table}

The Sewer-ML dataset is currently the largest open-source multi-label dataset in the field of sewer defect classification, containing over 1.3 million unique images collected by CCTV equipment. The images were extracted and annotated by sewer pipeline inspection professionals from 75,618 inspection videos between 2011 and 2019. The dataset covers 17 types of pipeline defects and normal images, with each sewer pipeline defect assigned a weight factor (Class Importance Weights, CIW) based on actual operational conditions, representing the socio-economic impact of the defect. Larger weights indicate more severe hazards and higher detection priority. The defect codes, descriptions, and weight factors are shown in Table \ref{tab:1}. The data covers main and branch pipelines of different materials, shapes, and sizes, with significant differences across pipelines, posing considerable challenges for classification models.

The Sewer-ML Dataset is divided into training, validation, and test sets by video segments, consisting of 60,356, 7,692, and 7,570 independent videos respectively. Images from different datasets come from different sewer pipeline videos, ensuring no duplicate images across subsets with uniform distribution of normal and defect samples. The overall dataset is divided into 80\% training set and 20\% validation test set (roughly equal split). The statistics of image numbers for each subset are shown in Table \ref{tab:2}. The ground truth labels for the test set are not publicly available, and model evaluation requires submission through the official website to ensure fair algorithm comparison.

\begin{table}[htbp]
\centering
\caption{Sewer-ML Dataset Multi-label Dataset Image Quantity Distribution}
\label{tab:2}
\begin{tabular}{l >{\raggedleft\arraybackslash}p{1.6cm} >{\raggedleft\arraybackslash}p{2.0cm} >{\raggedleft\arraybackslash}p{1.5cm}}
\toprule
\textbf{Type} & \textbf{Training} & \textbf{Validation} & \textbf{Test} \\
\midrule
Normal images & 552820 & 68681 & 69221 \\
Defect images & 487309 & 61365 & 60805 \\
\midrule
Total & 1040129 & 130046 & 130026 \\
\bottomrule
\end{tabular}
\end{table}

The Sewer-Capsule Dataset is collected based on pipeline capsule inspection equipment and is a single-label dataset used to verify algorithm effectiveness and generalization capability. Compared with v1, this dataset version has redefined and labeled defect categories, containing 5 classes including detachment, breakage, deformation, obstacles, and normal images, with 2,353 images in the training set and 1,177 in the validation set. Detailed category statistics are shown in Table \ref{tab:3}.

\begin{table}[htbp]
\centering
\caption{Sewer-Capsule Dataset Defect Categories and Quantity Distribution}
\label{tab:3}
\begin{tabular}{lrrrrr}
\toprule
\textbf{Type} & \textbf{Detachment} & \textbf{Breakage} & \textbf{Deformation} & \textbf{Obstacle} & \textbf{Normal} \\
\midrule
Training set & 620 & 371 & 227 & 197 & 938 \\
Validation set & 281 & 158 & 93 & 84 & 561 \\
\midrule
Total & 901 & 529 & 320 & 281 & 1499 \\
\bottomrule
\end{tabular}
\end{table}

\subsection{Evaluation Metrics}

Performance is evaluated using accuracy, precision, recall, $F1_{\text{Normal}}$, mean average precision (mAP), and the Sewer-ML benchmark's class-importance-weighted $F_2$ metric ($F2_{\text{CIW}}$). Because sewer defect categories have different operational consequences, $F2_{\text{CIW}}$ combines the classwise $F_2$ scores with the CIW factors listed in Table \ref{tab:1}:
\begin{linenomath*}
\begin{equation}
\text{$F2_{\text{CIW}}$} = \frac{\sum_{c=1}^{C} F_{2_c} \cdot \text{CIW}_c}{\sum_{c=1}^{C} \text{CIW}_c}
\end{equation}
\end{linenomath*}

where $F_{2_c}$ is the $F_2$ score for class $c$ and $\text{CIW}_c$ is its importance weight. $F1_{\text{Normal}}$ evaluates recognition of normal pipe images, and mAP summarizes performance across defect classes.

\section{Comprehensive Experiments on Sewer-ML Dataset}\label{sec:experiments}

\subsection{Comprehensive Multi-label Classification Evaluation}\label{original:6.1}

This experiment evaluates the performance of the multi-level vision Transformer-based Sewer-Transformer-ML model and the lightweight MobileNet-V3-based Sewer-MobileNet-ML model in multi-label classification of sewer defect images, and compares them with multiple SOTA models. Specifically, we trained four models: Sewer-Transformer-ML-Tiny, Sewer-Transformer-ML-Small, Sewer-Transformer-ML-Base, and Sewer-MobileNet-ML. Among them, Transformer-based models were trained for 300 epochs, while Sewer-MobileNet-ML was trained for 200 epochs. Table \ref{tab:5} shows the performance comparison of different methods on validation and test sets. The ground truth labels of the test set are not publicly available, requiring prediction results to be uploaded to the Sewer-ML Defect Classification Challenge official website for automatic evaluation to ensure leaderboard fairness. 

\begin{table}[htbp]
\centering
\caption{Comparison of Metrics Among Different Methods (Sewer-ML Dataset)}
\label{tab:5}
\begin{tabular}{lcccccccc}
\toprule
\multirow{2}{*}{Method} & \multicolumn{3}{c}{Validation dataset} & \multicolumn{2}{c}{Test dataset} & \multirow{2}{*}{size} \\ \cmidrule(lr){2-4} \cmidrule(lr){5-6}
 & mAP & $F2_{\text{CIW}}$ & $F1_{\text{Normal}}$ & $F2_{\text{CIW}}$ & $F1_{\text{Normal}}$ & \\ \midrule
\citeN{xie2019automatic} & - & 48.57 & 91.08 & 48.34 & 90.62 & 35M+35M \\
\citeN{chen2018intelligent} & - & 42.03 & 3.96 & 41.74 & 3.59 & 93 M \\
\citeN{hassan2019underground} & - & 13.14 & 0.00 & 12.94 & 0.00 & 220 M \\
\citeN{myrans2019automated} & - & 4.01 & 26.03 & 4.11 & 27.48 & 140 M \\
ResNet-101 \cite{he2016deep} & - & 53.26 & 79.55 & 53.21 & 78.57 & 162 M \\
KSSNet \cite{wang2020multi} & - & 54.42 & 80.60 & 54.55 & 79.29 & 173 M \\
TResNet-M \cite{ridnik2021tresnet} & - & 53.83 & 81.23 & 53.79 & 79.91 & 112 M \\
TResNet-L \cite{ridnik2021tresnet} & - & 54.63 & 81.22 & 54.75 & 79.88 & 205 M \\
TResNet-XL \cite{ridnik2021tresnet} & - & 54.42 & 81.81 & 54.24 & 80.42 & 290 M \\ \addlinespace
Sewer-Transformer-ML-Tiny & 62.97 & 60.1 & 86.35 & 60.28 & 85.07 & 107 M \\
\begin{tabular}[c]{@{}l@{}}Sewer-Transformer-ML-\\Small\end{tabular} & 63.19 & 65.59 & 89.16 & \textbf{66.63} & 88.24 & 188 M \\
Sewer-Transformer-ML-Base & \textbf{68.28} & \textbf{67.09} & \textbf{93.2} & 65.68 & \textbf{92.68} & 333 M \\
Sewer-MobileNet-ML & 63.09 & 66.01 & 90.1 & 65.73 & 89.53 & \textbf{17 M} \\ \bottomrule
\end{tabular}
\end{table}

\textbf{Leaderboard Performance.} Notably, on the official Sewer-ML Defect Classification Challenge test set, our Sewer-Transformer-ML-Base achieves 65.68\% $F2_{\text{CIW}}$ and 92.68\% $F1_{\text{Normal}}$, establishing a new state-of-the-art and ranking first on the public leaderboard. This represents a significant improvement of 7.6 percentage points over the second-best method (58.08\%), and surpasses the previous best published result (TResNet-XL, 54.24\%) by over 11 percentage points. Remarkably, even our lightweight Sewer-MobileNet-ML achieves 65.73\% $F2_{\text{CIW}}$ with merely 17M parameters (approximately 95\% parameter reduction compared to Sewer-Transformer-ML-Base and 94\% reduction compared to TResNet-XL), outperforming all existing CNN-based methods and validating that small models can achieve state-of-the-art accuracy in this domain.

\begin{figure}[htbp]
    \centering
    \includegraphics[width=0.48\linewidth]{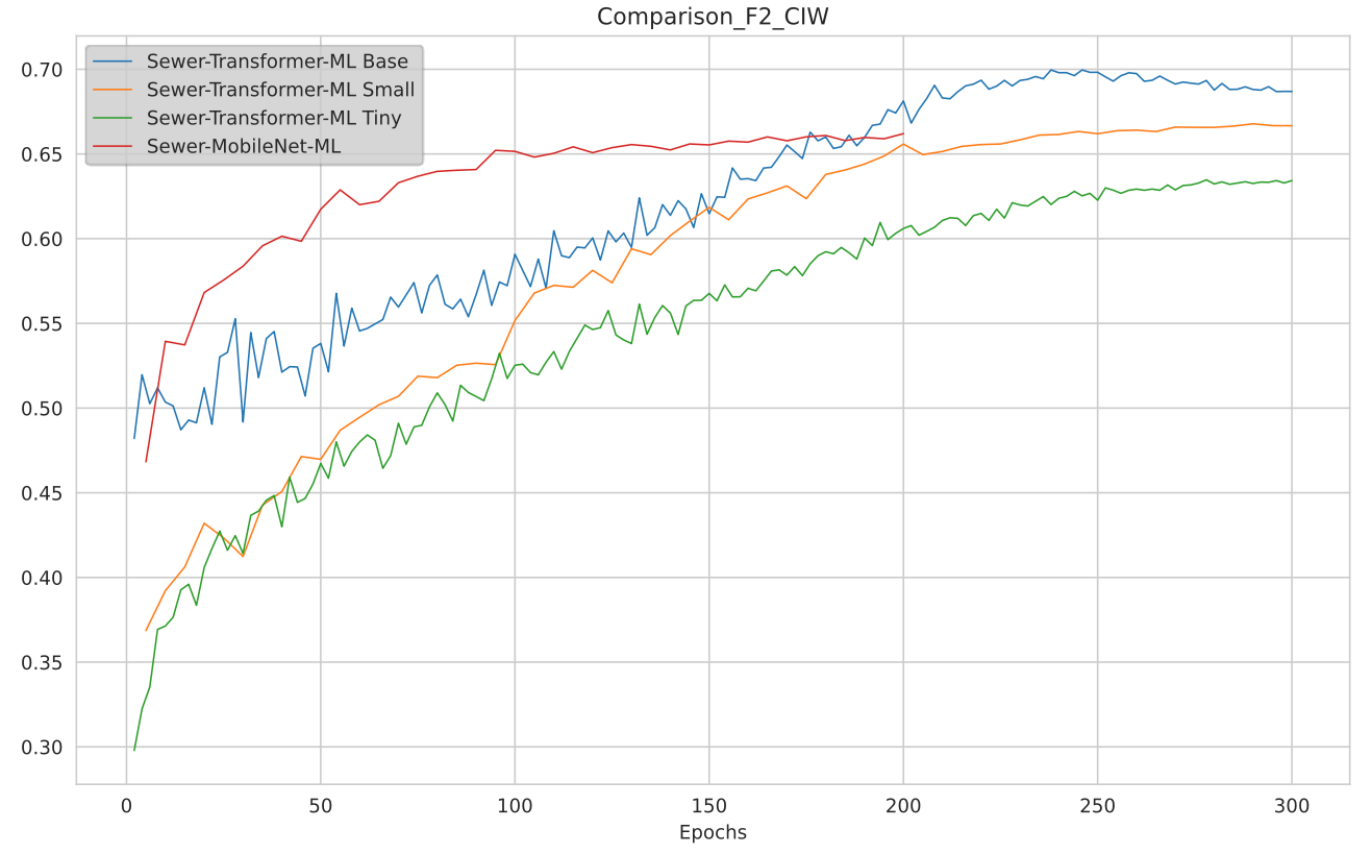}
    \caption{$F2_{\text{CIW}}$ Accuracy Curve Comparison}
    \label{fig:19}
\end{figure}

The models were evaluated on the complete validation set every 2 epochs. As shown by the $F2_{\text{CIW}}$ curves in Figure \ref{fig:19}, Sewer-MobileNet-ML stabilized after approximately 50 epochs, whereas the Sewer-Transformer-ML variants stabilized after about 150 epochs. Sewer-MobileNet-ML was trained for 200 epochs and the Transformer variants for 300 epochs.

All experiments were conducted on 4 A100 GPUs (40GB per card) using multi-GPU distributed training. The average training time for 200 epochs, including validation every 2 epochs, is reported in Table \ref{tab:6}; the lightweight model required substantially less training time than the Transformer variants.

\begin{table}[htbp]
\centering
\caption{Average Time Reference for Training 200 Epochs}
\label{tab:6}
\begin{tabular}{lc}
\toprule
\textbf{Method} & \textbf{Time (h)} \\ 
\midrule
Sewer-Transformer-ML-Tiny   & $\sim$ 40 h \\ \addlinespace[0.5em]
Sewer-Transformer-ML-Small  & $\sim$ 79 h \\ \addlinespace[0.5em]
Sewer-Transformer-ML-Base   & $\sim$ 110 h \\ \addlinespace[0.5em]
Sewer-MobileNet-ML          & $\sim$ 20 h \\ 
\bottomrule
\end{tabular}
\end{table}

\subsection{Multi-level Vision Transformer Feature Fusion Experiment}\label{original:6.2}

\begin{figure}[htbp]
    \centering
    \includegraphics[width=0.48\linewidth]{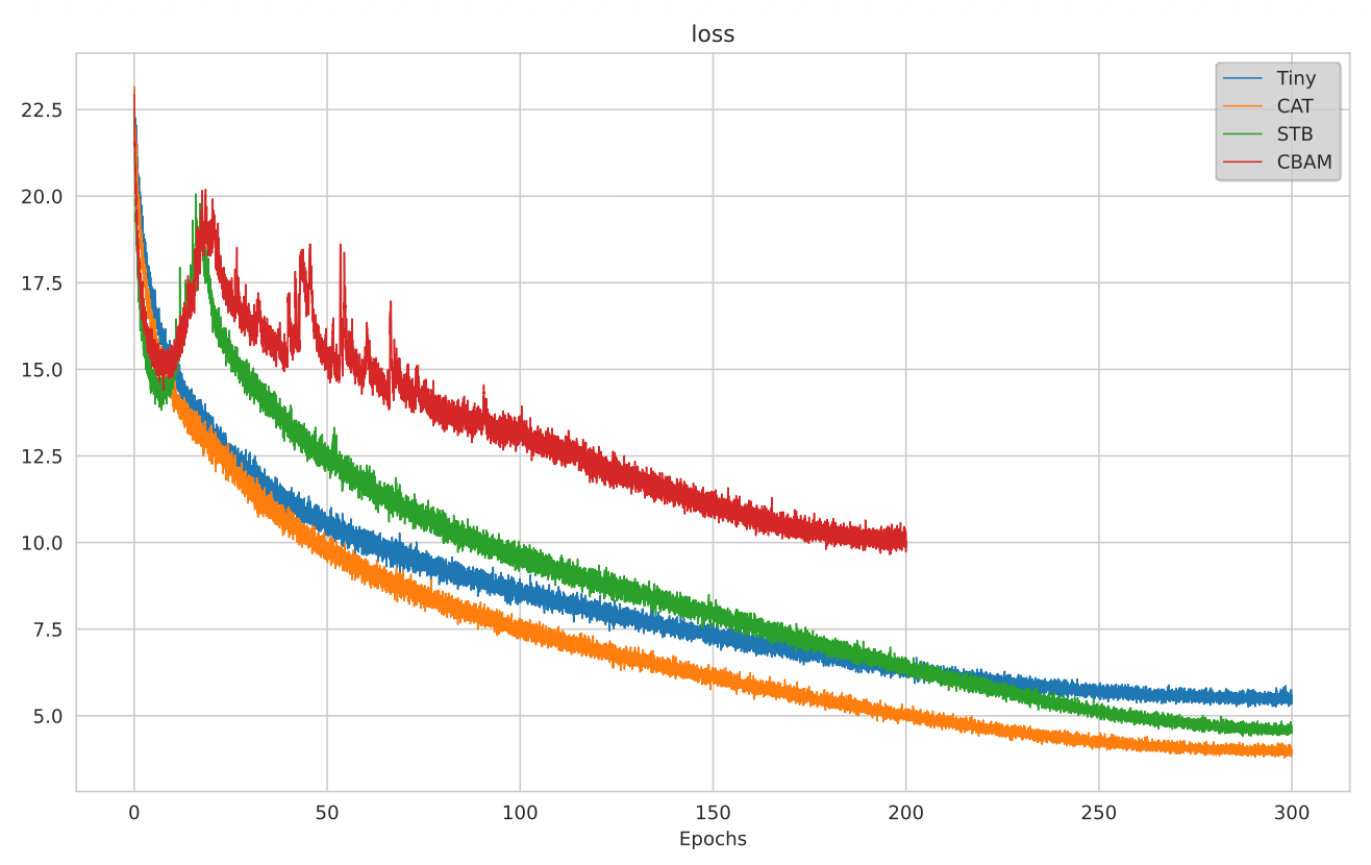}
    \caption{Loss curve comparison}
    \label{fig:21}
\end{figure}

\begin{figure}[htbp]
    \centering
    \begin{minipage}[t]{0.48\linewidth}
        \centering
        \includegraphics[width=\linewidth]{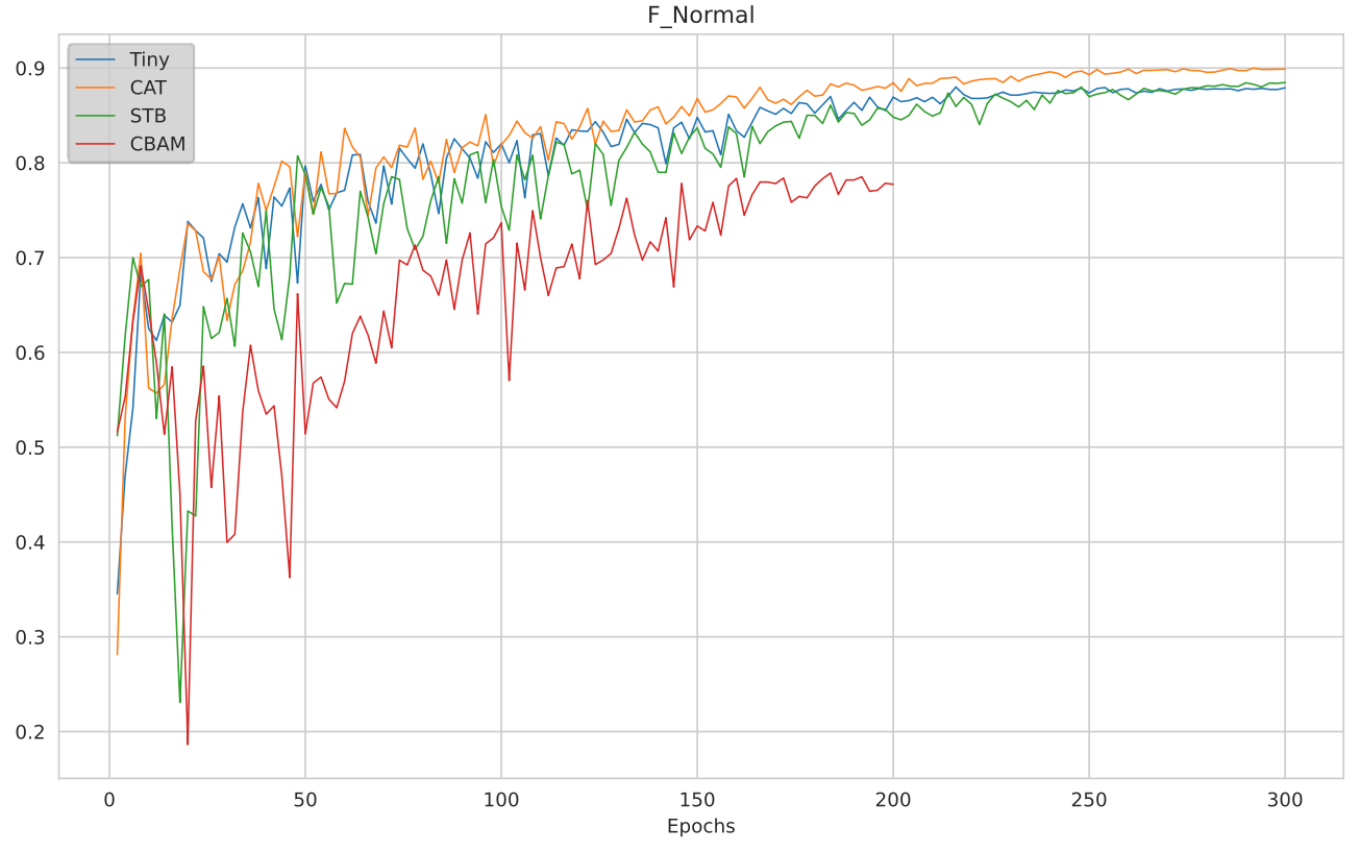}
        \captionof{figure}{F$_1$ Precision Curve Comparison}
        \label{fig:22}
    \end{minipage}\hfill
    \begin{minipage}[t]{0.48\linewidth}
        \centering
        \includegraphics[width=\linewidth]{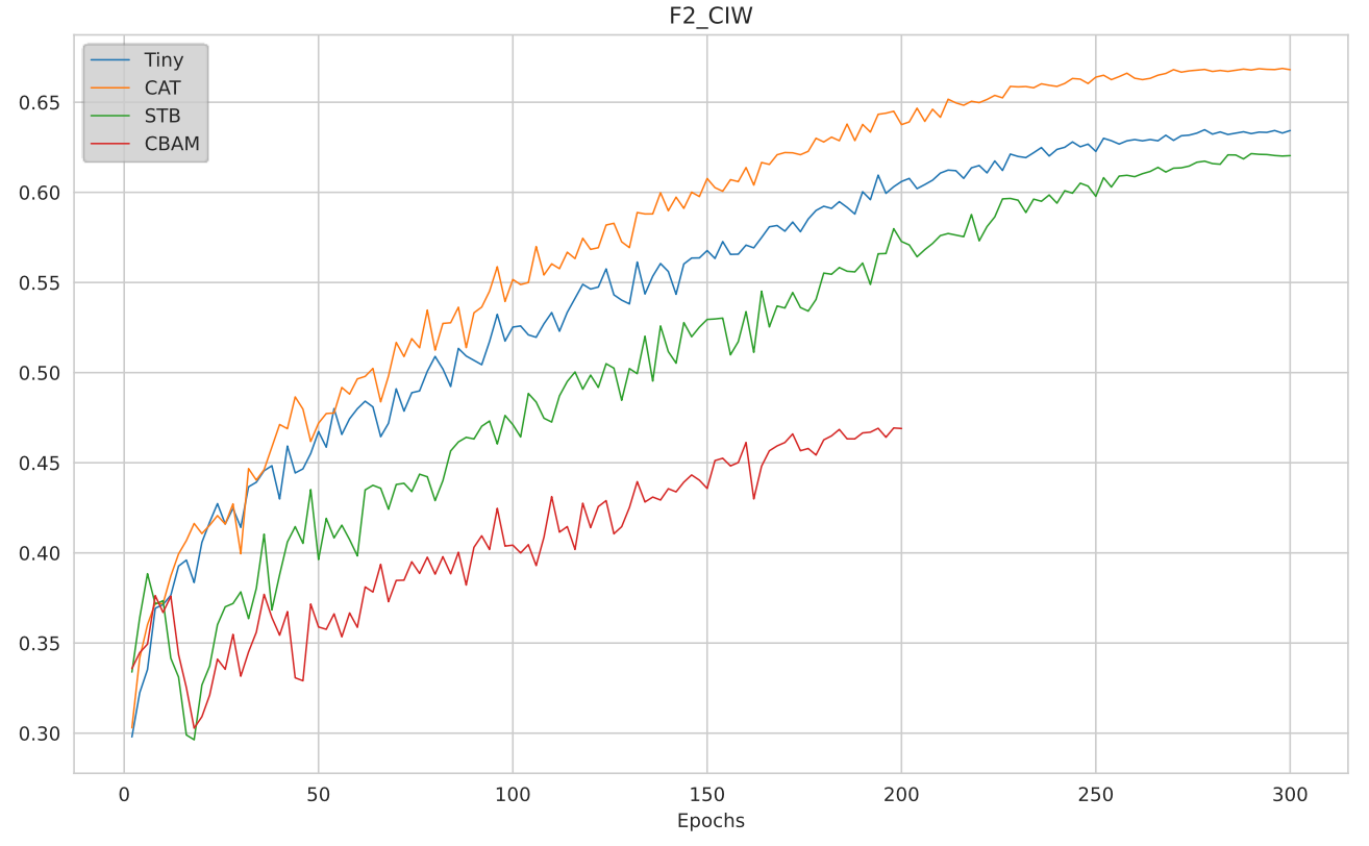}
        \captionof{figure}{$F2_{\text{CIW}}$ Accuracy Curve Comparison}
        \label{fig:23}
    \end{minipage}
\end{figure}

\begin{figure}[htbp]
    \centering
    \includegraphics[width=0.48\linewidth]{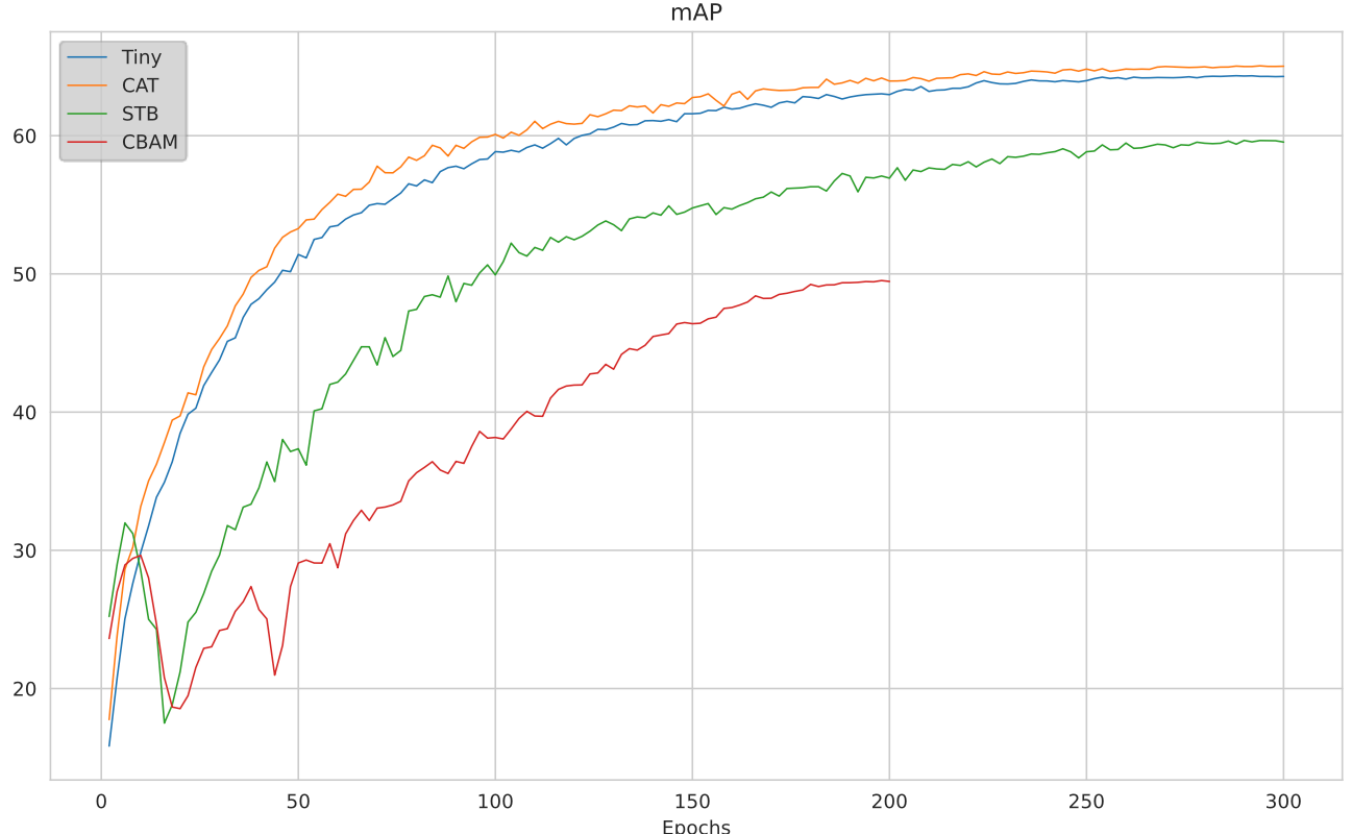}
    \caption{mAP accuracy curve comparison}
    \label{fig:24}
\end{figure}

Using the Sewer-Transformer-ML Tiny version as the base model, multi-level fusion experiments were conducted with qualitative and quantitative comparisons against other classical attention mechanism fusion strategies. The loss curves for each strategy are shown in Figure \ref{fig:21}, with  $F1_{\text{Normal}}$, $F2_{\text{CIW}}$, and mAP accuracy evaluation curves shown in Figure \ref{fig:22}, \ref{fig:23}, and \ref{fig:24} respectively. Among them, Tiny represents the baseline model without multi-level fusion strategy (using Sewer-Transformer-ML Tiny architecture, trained for 300 epochs), CAT represents multi-level feature direct stacking strategy (trained for 300 epochs), STB represents separated attention mechanism fusion strategy (trained for 300 epochs), and CBAM represents channel and spatial combined attention mechanism fusion strategy (trained for 200 epochs). During training, each metric on the validation set was evaluated every 2 epochs.

Experimental results show that in multi-label sewer defect classification, directly merging and stacking multi-level Transformer features (CAT) can improve model performance to some extent, but other attention mechanism-based fusion strategies do not further enhance performance. Analysis suggests this may be because features extracted at each stage of Transformer have already completed information screening and fusion through the self-attention mechanism, and introducing additional attention modules may instead lead to redundant computation. The CBAM strategy performed significantly weaker than other strategies during training, and all accuracy metrics stopped growing or even slightly decreased between epochs 170-200. To avoid overfitting, this strategy was terminated early after 200 epochs.

\subsection{Sewer-Mobile-TransNet Experiment}\label{original:6.3}

\begin{figure}[htbp]
    \centering
    \begin{minipage}[t]{0.48\linewidth}
        \centering
        \includegraphics[width=\linewidth]{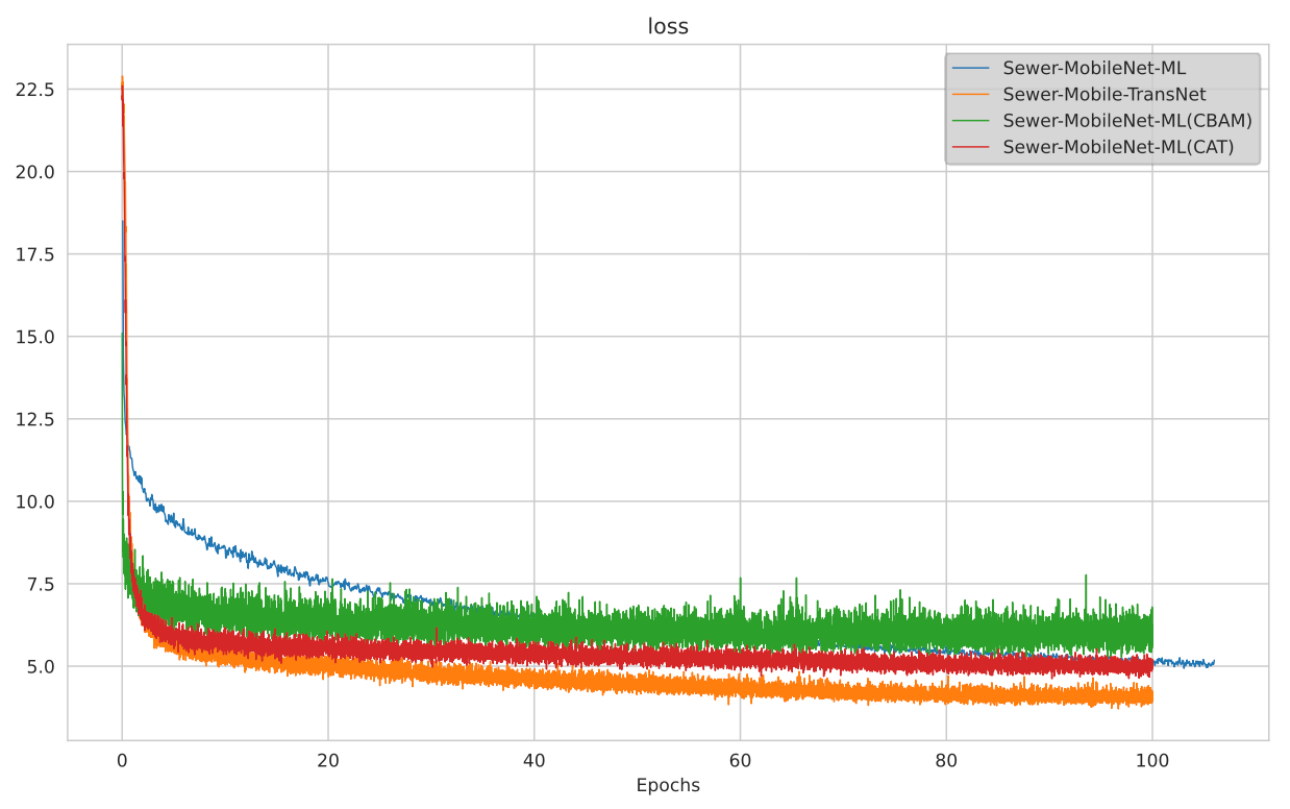}
        \captionof{figure}{Loss Curve Comparison}
        \label{fig:25}
    \end{minipage}\hfill
    \begin{minipage}[t]{0.48\linewidth}
        \centering
        \includegraphics[width=\linewidth]{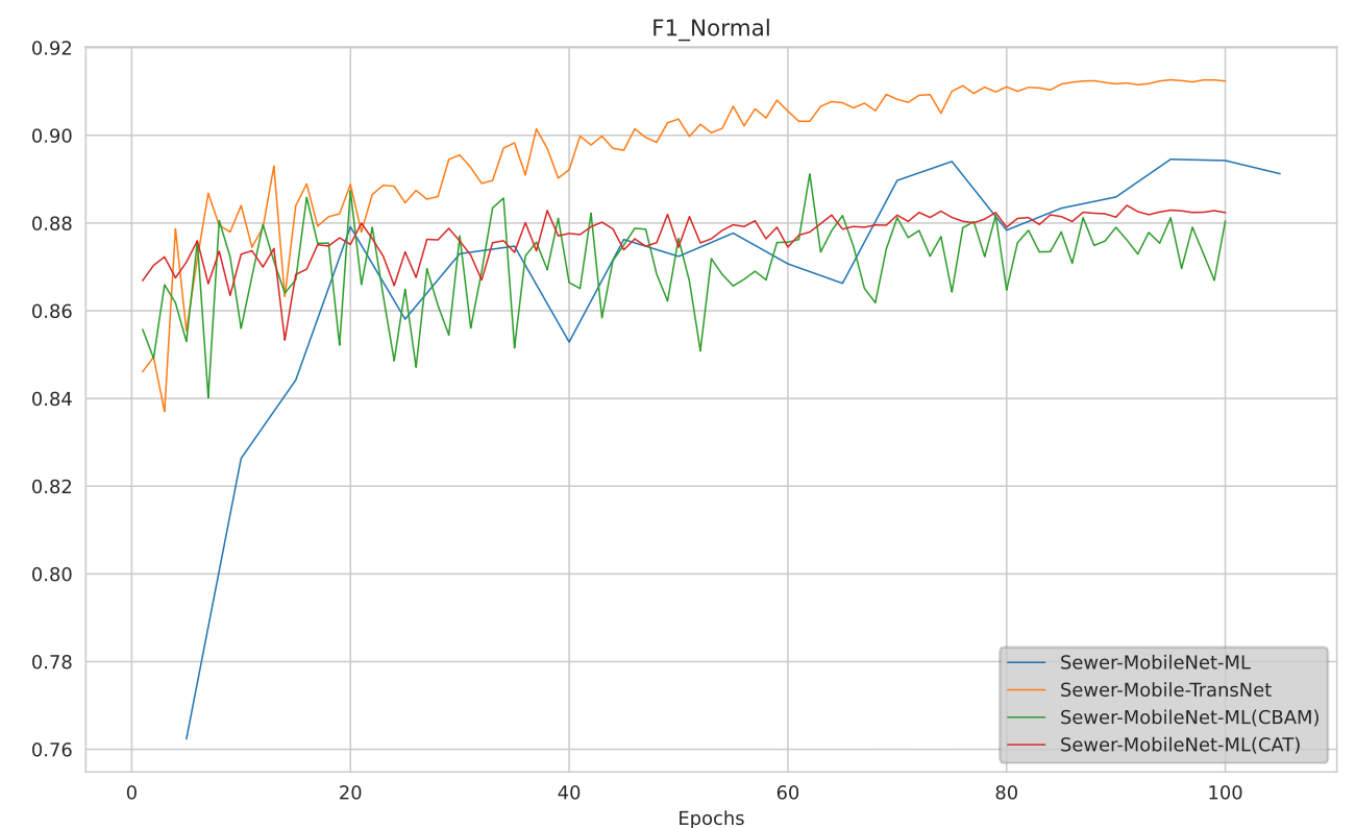}
        \captionof{figure}{F$_1$ Score Curve Comparison}
        \label{fig:26}
    \end{minipage}
\end{figure}

\begin{figure}[htbp]
    \centering
    \begin{minipage}[t]{0.48\linewidth}
        \centering
        \includegraphics[width=\linewidth]{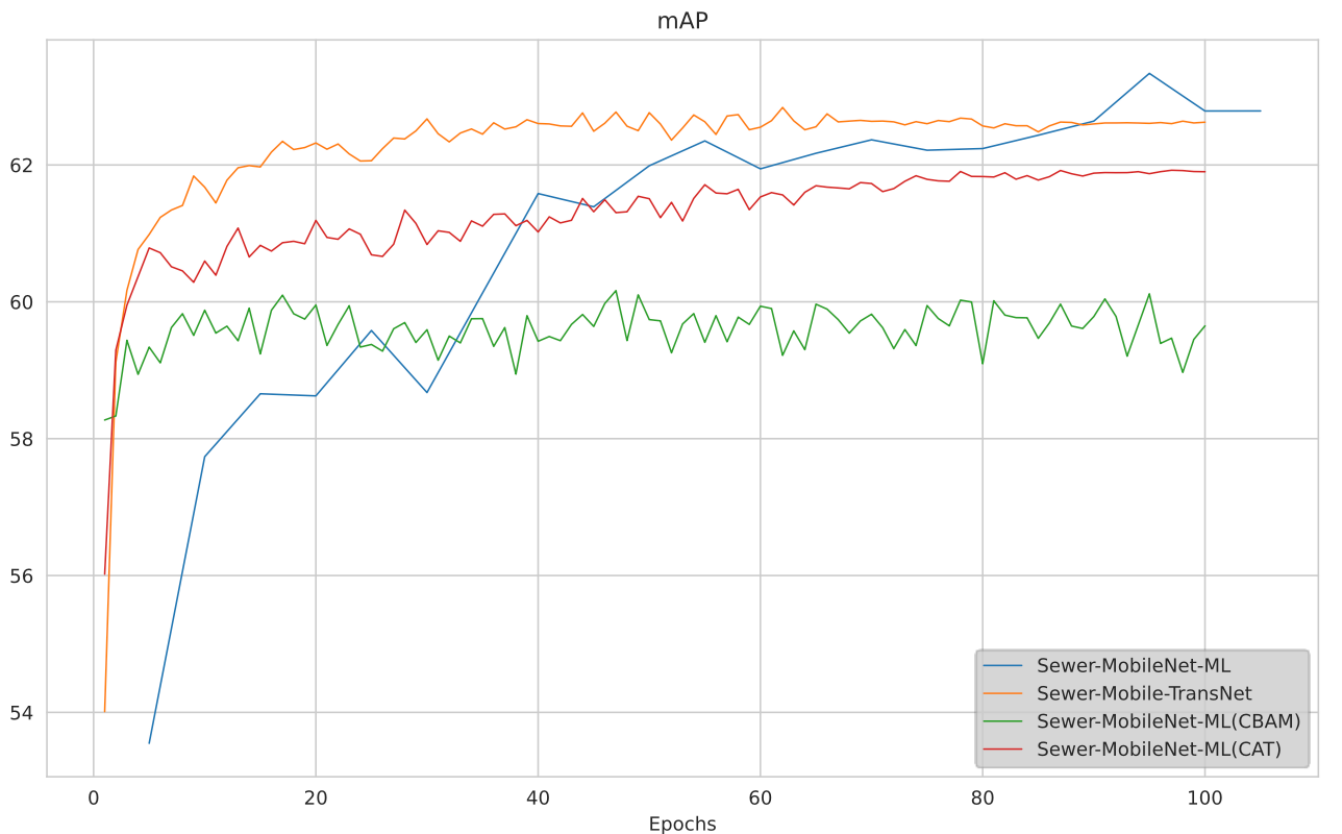}
        \captionof{figure}{mAP accuracy curve comparison}
        \label{fig:27}
    \end{minipage}\hfill
    \begin{minipage}[t]{0.48\linewidth}
        \centering
        \includegraphics[width=\linewidth]{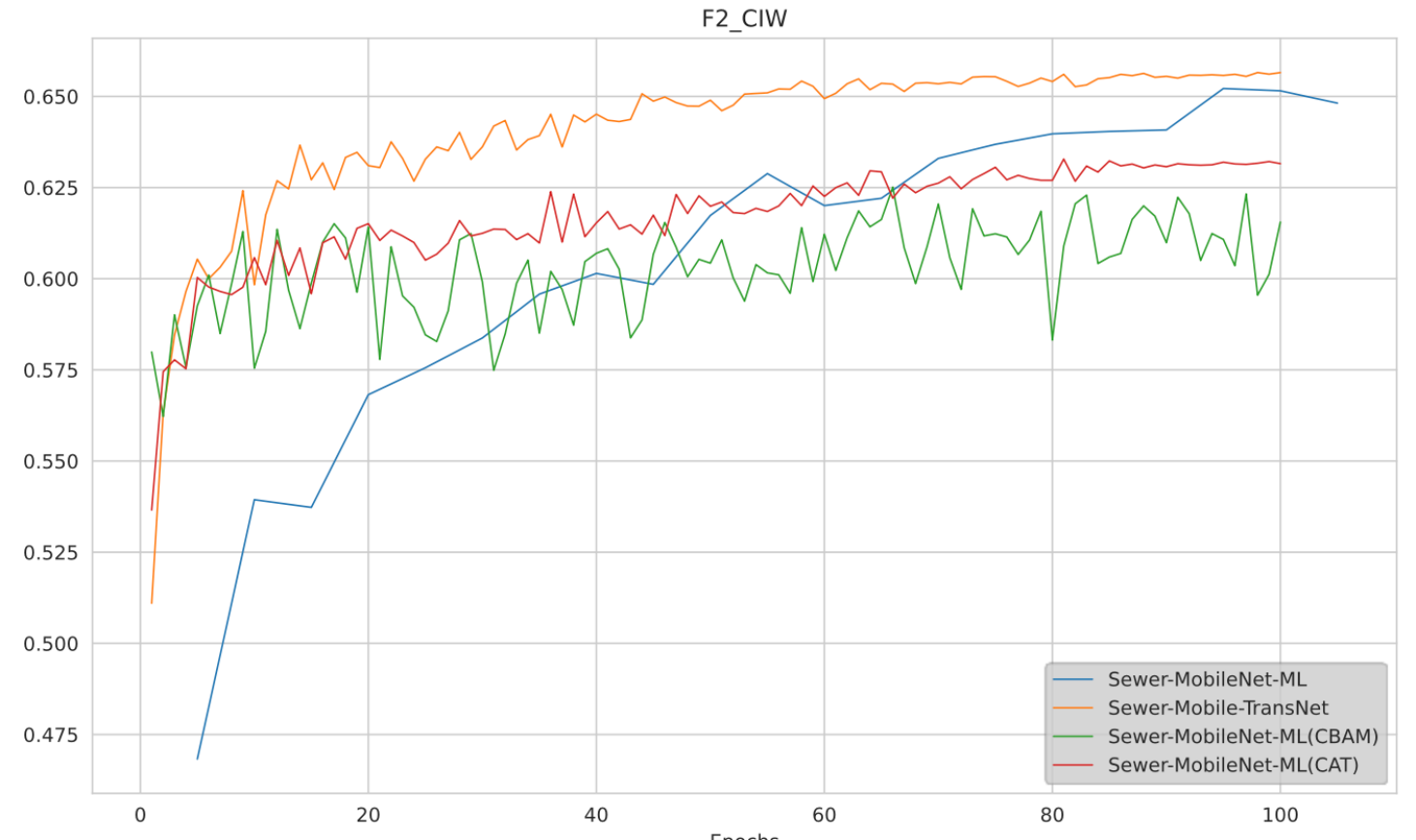}
        \captionof{figure}{$F2_{\text{CIW}}$ Accuracy Curve Comparison}
        \label{fig:28}
    \end{minipage}
\end{figure}

Lightweight design is a key factor for model practicality. Although the multi-level vision Transformer structure demonstrates excellent performance, its massive model weights (hundreds of MB) and long training time limit practical deployment. Therefore, we conducted lightweight experiments to reduce hardware costs, examine the potential for real-time detection, and shorten training cycles. Based on the Comprehensive Multi-label Classification Evaluation, Sewer-MobileNet-ML converges quickly, with all metrics stabilizing after approximately 50 epochs. In this experiment, Sewer-MobileNet-ML was trained for 105 epochs on the Sewer-ML dataset, with validation performed every 5 epochs. Sewer-Mobile-TransNet was trained for 100 epochs, with weights from Sewer-MobileNet-ML at epoch 90 fixed (only updating the subsequent fusion part). To contrast with the Multi-level Vision Transformer Feature Fusion experiment, CBAM and CAT multi-scale feature fusion strategies were also introduced. The model was evaluated on the validation set every epoch, with loss curves shown in Figure \ref{fig:25} and  $F1_{\text{Normal}}$, $F2_{\text{CIW}}$, and mAP evaluation curves shown in Figures \ref{fig:26}, \ref{fig:27}, and \ref{fig:28} respectively. 

Experimental results show that Sewer-Mobile-TransNet achieved the best performance across all evaluation metrics among lightweight architectures, further validating that small models with appropriate feature fusion can achieve state-of-the-art accuracy. Specifically, it demonstrates that even with approximately 95\% parameter reduction, the lightweight model maintains performance competitive with full-scale architectures, validating the effectiveness of our proposed multi-scale feature fusion strategy based on sliding-window multi-head self-attention.

\section{Sewer-Capsule Dataset Experiments}\label{sec:capsule}

Sewer-Capsule Dataset is a single-label dataset. The proposed methods can also be directly applied to single-label image classification tasks, with the main difference being the loss function during training. To validate the effectiveness of the method on pipeline capsule data and fully utilize the model weights obtained from pre-training on the large-scale Sewer-ML Dataset, we employ comprehensive transfer learning to generalize the model's classification capability to pipeline capsule data, improving the algorithm's robustness and adaptability, and reducing the deep learning model's dependency on training set data volume. Based on this, this section sets up two groups of comparative experiments. 

\subsection{Sewer-Capsule Image Defect Classification Experiment}\label{yinyong111}

The Sewer-Capsule Dataset training set contains 2,353 images, and the validation set contains 1,177 images. Models are trained for 40 epochs, with validation set accuracy evaluated every 2 epochs. Table \ref{tab:7} shows the optimal accuracy of Sewer-Transformer-ML (Tiny version), Sewer-Mobile-TransNet, and ResNet-50 models on the validation set. Experimental results indicate that all models perform well on the pipeline capsule dataset, with Sewer-Mobile-TransNet achieving 96.43\% accuracy, significantly outperforming Sewer-Transformer-ML (87.76\%) and ResNet-50 (87.51\%). This validates that small models achieve superior performance in transfer learning scenarios. Combined with the analysis in literature \cite{dosovitskiy2021image} of vision Transformer on the large-scale ImageNet dataset, this result further confirms that in sewer defect classification tasks, pure Transformer structures may not necessarily outperform convolutional network structures on small datasets. This is also consistent with the experimental results on Sewer-ML Dataset above, indicating that Transformer architectures are more suitable for large-scale image processing and analysis scenarios.

\begin{table}[htbp]
\centering
\caption{Sewer-Capsule Dataset Experimental Results}
\label{tab:7}
\begin{tabular}{lcccc}
\toprule
\textbf{Method} & \textbf{Acc (\%)} & \textbf{$F1_{\text{Normal}}$(\%)} & \textbf{Precision(\%)} & \textbf{Recall(\%)} \\
\midrule
Sewer-Mobile-TransNet & 96.43 & 94.90 & 93.53 & 96.72 \\
Sewer-Transformer-ML & 87.76 & 80.90 & 83.28 & 80.44 \\
ResNet-50 & 87.51 & 81.10 & 82.35 & 81.12 \\
\bottomrule
\end{tabular}
\end{table}

\subsection{Transfer Learning Experiment}

\begin{figure}[htbp]
    \centering
    \begin{minipage}[t]{0.48\linewidth}
        \centering
        \includegraphics[width=\linewidth]{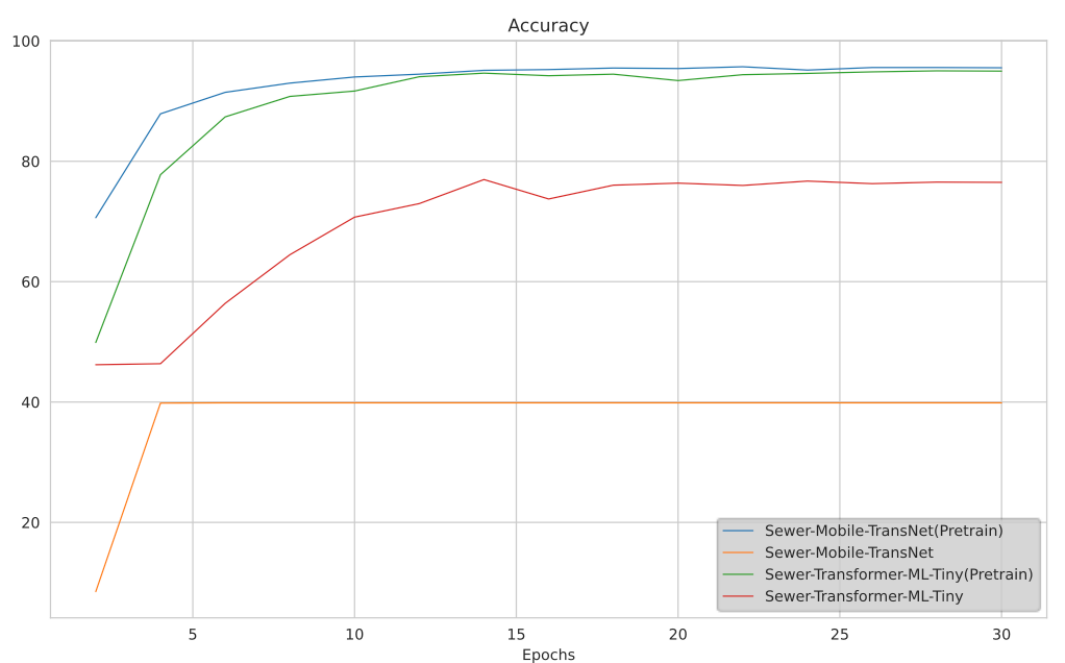}
        \captionof{figure}{Accuracy Curve Comparison (Sewer-Capsule Dataset)}
        \label{fig:29}
    \end{minipage}\hfill
    \begin{minipage}[t]{0.48\linewidth}
        \centering
        \includegraphics[width=\linewidth]{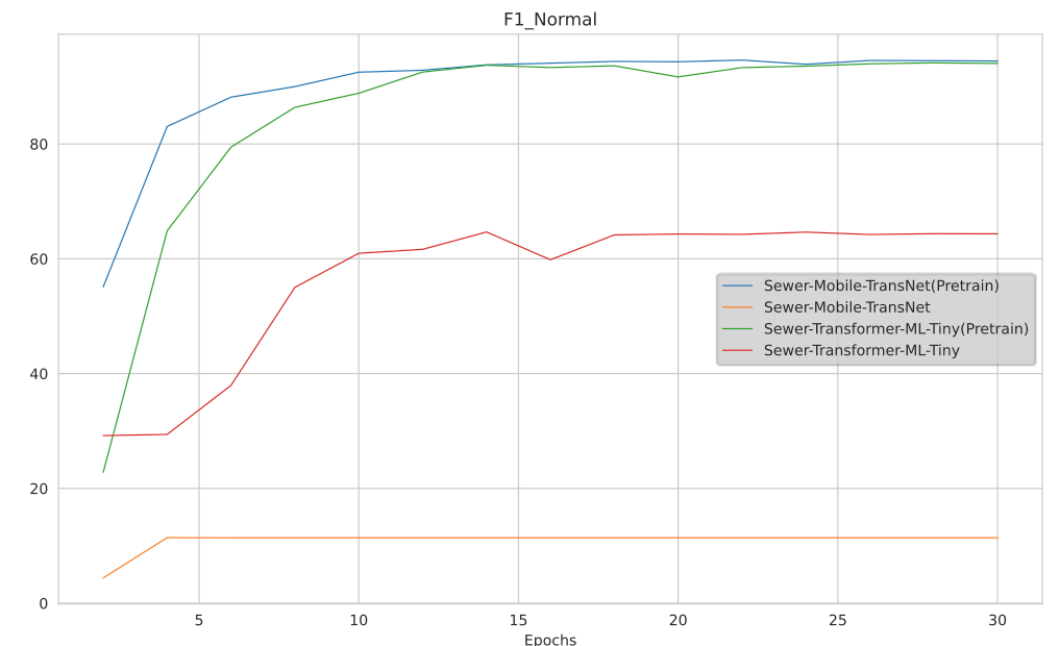}
        \captionof{figure}{F$_1$ Accuracy Curve Comparison (Sewer-Capsule Dataset)}
        \label{fig:30}
    \end{minipage}
\end{figure}

The training and validation set partition used in the Sewer-Capsule Image Defect Classification experiment, although helpful for obtaining good generalization performance, has high costs for acquiring high-quality ground-truth labels in practical applications. We expect to achieve better results with less data. To this end, we swapped the training and validation sets from that experiment, creating a setup with only 1,177 images in the training set and 2,353 images in the validation set. Simultaneously, each model has two groups of comparative experiments: using random initialization and using pre-trained weights based on the Sewer-ML dataset, respectively, to improve model classification performance through transfer learning. Models are trained for 30 epochs, with the validation set evaluated every 2 epochs. The Acc and $F1_{\text{Normal}}$ metric trends are shown in Figures \ref{fig:29} and \ref{fig:30}, respectively (Pretrain in the figures indicates model weights obtained from pre-training on the Sewer-ML dataset). Experimental results show that after using pre-trained weights, all accuracy metrics are significantly improved, validating the model's strong cross-domain generalization capability even with limited data.

\section{Discussion and Analysis}\label{sec:discussion}

Experimental results demonstrate that both Sewer-Transformer-ML and lightweight Sewer-MobileNet-ML models exhibit excellent performance in multi-label classification of sewer defect images. Among them, the Sewer-Transformer-ML Base model achieved the best results on the Sewer-ML dataset and ranked first on the Sewer Defect Classification Challenge leaderboard, significantly outperforming the second-best method by 7.6 percentage points (65.68\% vs. 58.08\%). The Sewer-Transformer-ML Small and Tiny versions also surpassed the performance of classic convolutional neural network models such as TResNet-XL and ResNet-101. This result confirms that a purely Transformer-based architecture can effectively capture dependencies between labels in multi-label classification tasks, achieving better classification results than CNN structures. From the perspective of trade-offs between model size and performance, although larger parameter models theoretically have stronger performance, the Small and Tiny versions, with nearly half the parameters reduced, only see a 1 to 6 percentage point drop in core evaluation metrics, demonstrating good cost-effectiveness.

Although the Sewer-Transformer-ML Small and Tiny versions have been optimized in model size, they still struggle to meet deployment requirements for mobile or IoT terminals. In contrast, the lightweight model Sewer-MobileNet-ML, while achieving approximately 95\% parameter reduction (only about one twentieth of Sewer-Transformer-ML), achieved state-of-the-art accuracy (65.73\% $F2_{\text{CIW}}$) among current end-to-end CNN methods and set a new record for lightweight architectures. Its core accuracy metrics are already very close to the Sewer-Transformer-ML Base version, with some metrics even surpassing the Small and Tiny variants, validating that small models can achieve superior performance.

Through comprehensive and systematic ablation experiments, to further explore the characteristics of Transformer multi-level features, we designed three fusion strategies: direct feature concatenation (CAT), separated attention mechanism (STB), and channel-spatial attention mechanism (CBAM). Comparative results show that directly stacking Transformer multi-level features can effectively improve model performance, while additionally introducing attention modules does not bring further improvement. Analyzing the reasons, features extracted at each stage of Transformer have already completed information screening and fusion through self-attention mechanisms. Applying attention mechanisms again may introduce redundant computation and instead limit performance improvement.

We conducted thorough experiments on the Sewer-Mobile-TransNet structure to further explore the differences between CNN multi-scale features and Transformer multi-level features. Similar to the Multi-level Vision Transformer Feature Fusion experiment, we performed fusion analysis on multi-scale features extracted by MobileNetV3 and compared two strategies: direct concatenation and channel-spatial attention mechanism. The experiments found that using multi-head self-attention mechanism to fuse CNN multi-scale features can effectively improve model performance.

Comprehensive comparison of the Multi-level Vision Transformer Feature Fusion and Sewer-Mobile-TransNet experiments shows that multi-level features extracted by vision Transformer and multi-scale features extracted by CNN have essential differences. This also indicates that conventional fusion strategies for CNN may not be effective for multi-level features of vision Transformer. CNN multi-scale features are more suitable for fusion through attention mechanisms to improve model performance, while direct concatenation of Transformer multi-level features yields better results. This discovery provides important reference basis for structural design of deep learning image classification models.

Through comprehensive transfer learning experiments, we validated the defect classification capability of the proposed method on the pipeline capsule dataset. Sewer-Mobile-TransNet achieved 96.43\% accuracy under the standard data split with 2,353 training images. When the training set was reduced to 1,177 images, transferring pre-trained weights from the large-scale Sewer-ML dataset consistently improved model performance compared with random initialization. These results demonstrate that useful sewer defect representations can be transferred across inspection platforms, reducing dependence on target-domain annotated data and enhancing the engineering practicality of the method.

\subsection{Limitations and Future Work}\label{sec:limitations}

Despite the demonstrated performance, this study has several limitations. First, the models were evaluated primarily on the Sewer-ML and Sewer-Capsule data sets. The Sewer-Capsule data set is relatively small and was collected using a specific inspection platform; therefore, the reported results may not fully represent generalization across different cities, pipe materials, defect standards, imaging conditions, and robotic systems. Second, the ground-truth labels of the Sewer-ML test set are not publicly available, and test performance must be obtained through the official evaluation server. This restriction prevents a more detailed independent analysis of class-specific errors and failure cases on the test set. Third, lightweight performance was assessed mainly through parameter count, classification accuracy, and training time. Inference latency, memory consumption, and energy use have not yet been systematically evaluated on mobile or embedded hardware. Consequently, the results demonstrate the potential for edge deployment rather than a completed field deployment. Future research will focus on cross-region and cross-device evaluation, hardware-level deployment tests, few-shot and weakly supervised learning, and extensions toward defect localization, severity assessment, and longitudinal condition analysis.

\section{Conclusion}\label{sec:conclusion}

This study addressed multi-label defect classification in urban sewer inspection images by developing Sewer-Transformer-ML, a hierarchical vision Transformer with multi-level feature fusion. Two lightweight architectures, Sewer-MobileNet-ML and Sewer-Mobile-TransNet, were further investigated to examine the balance among classification performance, model complexity, and engineering application potential.

On the Sewer-ML test set, Sewer-Transformer-ML-Base achieved an $F2_{\text{CIW}}$ of 65.68\% and an $F1_{\text{Normal}}$ of 92.68\%, ranking first on the public challenge leaderboard and exceeding the second-ranked method by 7.6 percentage points in the primary metric. Sewer-MobileNet-ML contained only 17 M parameters, approximately 95\% fewer than the 333 M-parameter base model, while achieving an $F2_{\text{CIW}}$ of 65.73\%. These results indicate that large model size is not the only route to high-performance sewer defect classification and that lightweight architectures can also achieve competitive performance.

The feature-fusion experiments further demonstrated that the appropriate fusion strategy depends on the underlying network architecture. Direct concatenation was more effective for multi-level Transformer features, which had already undergone self-attention-based information interaction, whereas attention-based fusion provided greater benefits for multiscale CNN features. This distinction offers a transferable computational design insight for feature-fusion models used in civil infrastructure image analysis.

Under the standard Sewer-Capsule data split, Sewer-Mobile-TransNet achieved 96.43\% classification accuracy. In the reduced-data experiment using 1,177 training images, transferring pretrained knowledge from Sewer-ML consistently improved model performance. Overall, the proposed framework can support automated analysis of large-scale CCTV inspection data and shows potential for adaptation to emerging robotic inspection platforms. Further cross-region evaluation and testing on operational inspection hardware are required to establish field generalizability, computational efficiency, and long-term deployment reliability.

\appendix
\section{Experimental Training Settings Summary}\label{app:training_settings}

\subsection{Sewer-ML Model Series Training Configurations}\label{app:sub:sewer_ml}

The detailed training configurations for the Sewer-Transformer-ML series models are presented below:

\begin{itemize}
    \item \textbf{Sewer-Transformer-ML Series Models:} Trained on 4 Nvidia A100 GPUs, using the AdamW optimizer. The weight decay coefficient is set to 0.05. The learning rate is scaled with the total batch size according to the linear scaling rule:
          \[
              \text{lr}_{\text{actual}} = 5 \times 10^{-4} \times \frac{\text{Batch\_Size} \times \text{GPU\_NUM}}{512}
          \]
          where \texttt{GPU\_NUM} (here, 4) denotes the number of GPUs used. The per-GPU batch sizes for different model variants are as follows: 320 for Tiny, 196 for Small, and 144 for Base.

    \item \textbf{Sewer-MobileNet-ML:} Trained on 4 Nvidia A100 GPUs, using the RMSprop optimizer with momentum set to 0.9 and weight decay coefficient set to $1 \times 10^{-5}$. The learning rate follows the same scaling rule as above. The per-GPU batch size is set to 512.
\end{itemize}

\subsection{Additional Experiment Configurations}\label{app:sub:additional_exp}

\begin{itemize}
    \item The Multi-level Vision Transformer Feature Fusion and Sewer-Mobile-TransNet experiments in the main text used the same training configurations as the \textbf{Sewer-Transformer-ML} and \textbf{Sewer-MobileNet-ML} models described in Appendix \ref{app:sub:sewer_ml}, respectively.
\end{itemize}

\subsection{Comparative Study Training Configurations}\label{app:sub:comparative}

\begin{itemize}
    \item \textbf{Sewer-Mobile-TransNet:} Used the same optimizer and hyperparameter setup (RMSprop, etc.) as the Sewer-MobileNet-ML model in Appendix \ref{app:sub:sewer_ml}, but with a batch size of 64 and trained on a single GPU.

    \item \textbf{Sewer-Transformer-ML (Tiny):} Used the same optimizer and hyperparameter setup (AdamW, etc.) as the Sewer-Transformer-ML model in Appendix \ref{app:sub:sewer_ml}, with a batch size of 64 and trained on a single GPU.

    \item \textbf{ResNet-50 (Baseline):} Optimizer: SGD with Nesterov momentum (0.9). Initial learning rate: 0.003. Weight decay: 0.001. Batch size: 64. Trained on a single GPU.
\end{itemize}

\sAppendix{Data Availability Statement}

Some or all data, models, or code that support the findings of this study are available from the corresponding author upon reasonable request.

\sAppendix{Acknowledgments}

This work was supported by the National Natural Science Foundation of China (Grant No. 62272313).

\bibliography{autosam}

\end{document}